\documentclass[copyright,creativecommons]{eptcs}
\providecommand{\event}{LoCoCo 2011} 

\usepackage{graphicx}
\usepackage{latexsym}
\usepackage{array}
\usepackage{color}

\usepackage{alltt}
\usepackage[T1]{fontenc}

\renewcommand\texttt[1]{\mbox{\fontfamily{cmtt}\fontsize{10}{10}\selectfont#1}}

\usepackage{amsmath}
\usepackage{amsfonts}
\usepackage{amssymb}

\newcommand{\PS}{{\bf{P_S}}}
\newcommand{\PR}{{\bf{P_R}}}
\newcommand{\R}{{\bf{R}}}
\newcommand{\Sol}{{\bf{S}}}
\newcommand{\T}{{\bf{T}}}
\newcommand{\REQ}{{\bf{REQ}}}

\newcommand{\LS}{{\bf{L}}}

\newcommand{\HM}{\mathcal{H}}

\newcommand{\reuse}{\mathtt{reuse}}

\newtheorem{definition}{Definition}

\begin{document}

\title{(Re)configuration based on model generation\thanks{This work has been developed within the scope of the project RECONCILE (reconciling legacy instances with changed ontologies) and was funded by FFG FIT-IT (grant number 825071).}
}

\author{
Gerhard Friedrich \\ Anna Ryabokon
\institute{Alpen-Adria Universit\"at, 
Klagenfurt, Austria}
\email{firstname.lastname@aau.at}
\and
Andreas A.\ Falkner \quad Alois Haselb\"ock \\ 
Gottfried Schenner \quad Herwig Schreiner
\institute{Siemens AG Österreich, Vienna, Austria}
\email{firstname.\{middleinitial.\}lastname@siemens.com}
}
\def\titlerunning{(Re)configuration based on model generation}
\def\authorrunning{G.\ Friedrich, A.\ Ryabokon, A.A.\ Falkner, \\ A.\ Haselb\"ock, G.\ Schenner, H.\ Schreiner}

\maketitle

\begin{abstract}
Reconfiguration is an important activity for companies selling configurable products or services which have a long life time. However, identification of a set of required changes in a legacy configuration is a hard problem, since even small changes in the requirements might imply significant modifications.
In this paper we show a solution based on answer set programming, which is a logic-based knowledge representation formalism well suited for a compact description of (re)configuration problems. Its applicability is demonstrated on simple abstractions of several real-world scenarios. 
The evaluation of our solution on a set of benchmark instances derived from commercial (re)configuration problems shows its practical applicability. 
\end{abstract}

\section{Introduction} 

Reconfiguration is an important task in the after-sale life-cycle of configurable products and services, because requirements for these products and services are changing in parallel with the customers' business~\cite{Manhart2005,Falkner10}. In order to keep a product or a service up-to-date a re-engineering organization has to decide which modifications should be introduced to an existing configuration such that the new requirements are satisfied but change costs are minimized. 

Following the knowledge based configuration approach, we formulate reconfiguration problem instances as extensions of declaratively defined configuration problem instances where configurations are represented by facts and requirements are expressed by logical descriptions. These requirements may be partitioned into customer requirements and system specific configuration requirements. A configuration is simply defined as a subset of a logical model of the requirements. Informally, a reconfiguration problem instance is generated by an adaption of the requirements  resulting in a new set of  requirements and therefore a new instance of a configuration problem is formulated. Subsequently, given legacy configurations have to be adapted to configurations for the new requirements.  In our approach, the knowledge base comprises two parts: (1) the description of the new configuration problem instance, where all valid configurations are specified by the set of adapted requirements and (2) transformation knowledge regarding reuse and deletion of parts of a legacy configuration. Technically speaking this is a mapping from facts describing the legacy configuration to facts in the ontology of the new configuration problem instance. For generating a reconfiguration the problem solver has to decide which parts of the legacy configuration are either reused or deleted and which new parts have to be created.  

We introduce general definitions for (re)configuration problems employing 
Herbrand-models of logical descriptions.
Because of the remarkable advances of answer set programming (ASP)~\cite{
Simons2002,Leone2006,Gebser2010a} we base our implementation on this reasoning framework. 
ASP was first applied to configuration problems by~\cite{Soininen2001}. In 
particular, we provide modeling patterns for (re)configuration which allow 
the generation of optimized reconfigurations exploiting standard ASP solvers. 
Finally, our evaluation shows that the proposed method solves reconfiguration 
problem instances which are practically interesting for industrial 
applications.

\section{Example} \label{sec:example}

Let us exemplify different (re)configuration scenarios on a problem which is a simple abstraction of several configuration problems occurring in practice, i.e.\ entities may be contained in other entities but some restrictions must be fulfilled. We employ the ontology comprising the concepts \emph{person}, \emph{thing}, \emph{cabinet}, and \emph{room} where persons are related to things, things are related to cabinets, cabinets are related to rooms, and rooms are related to persons. 

As input to the configuration problem an ownership relation between persons and things is provided. We call this input a customer requirement since it reflects the individual needs of a customer using a configuration system whereas configuration requirements specify the properties of the system to be configured. Each person can own any number of things but each thing belongs to only one person. The problem is to place these things into cabinets and the cabinets into rooms of a house such that the following configuration requirements are fulfilled:
(1) each thing must be stored in a cabinet;
(2) a cabinet can contain at most 5 things;
(3) every cabinet must be placed in a room;
(4) a room can contain at most 4 cabinets;
(5) a person can own any number of rooms;
(6) each room belongs to a person;
and (7) a room may only contain cabinets storing things of the owner of the room.

\begin{figure}[b]
\begin{minipage}[t]{0.45\linewidth}
\centering
\includegraphics[width=\linewidth]{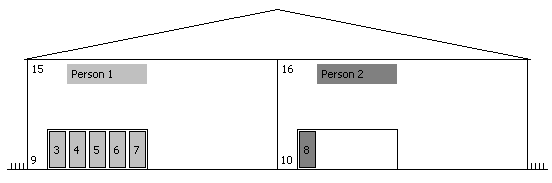}
\caption{Initial configuration}
\label{figconfig}
\end{minipage}
\hspace{0.5cm}
\begin{minipage}[t]{0.45\linewidth}
\centering
\includegraphics[width=\linewidth]{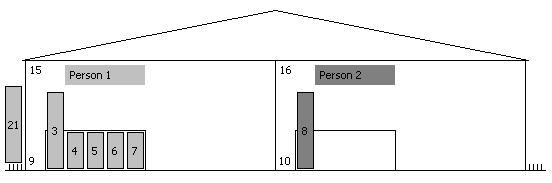}
\caption{Reconfiguration -- initial state}
\label{figreconfiginitial}
\end{minipage}
\end{figure}

In our simple example we only consider configuration of one house and represent all individuals using unique integer identifiers.
Informally, a configuration is every instantiation of the relations which satisfies all requirements. 
Let a sample house problem instance include two persons. The first person owns five things numbered 3 to 7 and the second person owns one thing 8. A solution for this problem instance is shown in Figure~\ref{figconfig}, including rooms 15 and 16 with cabinets 9 and 10.
Reconfiguration is necessary, whenever the customer requirements or configuration requirements are changed. For instance, it must be differentiated between long and short things with the following additional requirements: 
(8) a cabinet is either small or high;
(9) a long thing can only be put into a high cabinet;
(10) a small cabinet occupies 1 and a high cabinet 2 of 4 slots available in a room;
and (11) all legacy cabinets are small.

The customer requirements, in this case, define for each thing if it is long or short. For instance, things 3 and 8 are long; all others are short. Moreover, the first person gets an additional long thing 21. 
The changes to the legacy configuration are summarized in Figure~\ref{figreconfiginitial} showing an inconsistent configuration, where thing 21 is not placed in any of the cabinets, and cabinets 9 and 10 are too small for things 3 and 8.
\begin{figure}[tb]
\begin{minipage}[b]{0.45\linewidth}
\centering
\includegraphics[width=\linewidth]{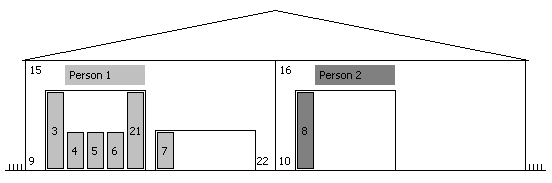}
\caption{Reconfiguration solution 1}
\label{figreconfigsol1}
\end{minipage}
\hspace{0.5cm}
\begin{minipage}[b]{0.45\linewidth}
\centering
\includegraphics[width=\linewidth]{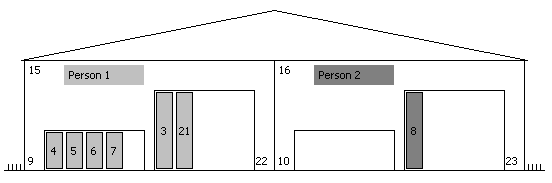}
\caption{Reconfiguration solution 2}
\label{figreconfigsol2}
\end{minipage}
\end{figure}
To obtain a solution which is shown in Figure~\ref{figreconfigsol1} the reconfiguration process changes the size of cabinets 9 and 10 to high and puts the new thing 21 into cabinet 9. A new small cabinet 22 is created for thing 7.

In our reconfiguration process every modification to the existing configuration, i.e. reusing/deleting/ creating individuals and their relations, is associated with some cost. Therefore, the reconfiguration problem is to find a consistent configuration by removing the inconsistencies and minimizing the costs involved. Different solutions will be 
found depending on the given modification costs.
If, for example, the costs for adding a new high cabinet are less
than the cost for changing an existing small cabinet into a high cabinet, then the previous solution should be rejected as its costs are too high. One of the solutions with less reconfiguration costs (see Figure~\ref{figreconfigsol2}) includes two new cabinets 22 and 23. 
Furthermore, cabinet 10 is not removed because it's cheaper to keep the cabinet than to delete it. 

\section{Short overview of answer set programming} \label{sec:review}

Before we introduce (re)configuration problems we give a brief overview of ASP. ASP is based on a decidable fragment of first-order logic enhanced with default negation and aggregation. A detailed discussion of ASP can be found in~\cite{Leone2006,Gebser2010a}.
In order to simplify our presentation we avoid default negation and employ a restricted form of aggregation called \emph{cardinality} 
constraint~\cite{Soininen2001,Simons2002,Gebser2010a}. Cardinality constraints are of the form $l \{a_1, \ldots, a_n\} u$
 where $a_i$ are atoms and $l, u$ are integers specifying lower and upper bounds. By using cardinality constraints in the consequent of a rule, choices can be expressed, i.e.\ from the atoms $a_i$ at least $l$ but at most $u$  must be true. Furthermore, the answer set semantic restricts the set of classical Herbrand models. Only those Herbrand models are accepted where every atom is either a fact or justified by a non-circular application of rules. For instance, \ $0 \{a,b\} 1 \leftarrow c$ is satisfied by $\{a\}$ but $\{a\}$ is not justified. However, if we add the fact $c$ to the knowledge base, $\{c\}$, $\{c, a\}$, and $\{c,b\}$ are stable models, i.e.\ Herbrand models accepted by the answer set semantic of~\cite{Simons2002}. 
ASP allows logical variables under syntactical restrictions which assure the elimination of all variables by ground terms, i.e.\ during grounding of a knowledge-base a rule containing variables is replaced by a set of ground rules. 

 For a succinct specification of facts in our example we use so-called intervals, e.g.\ $\mathtt{person(1..2).}$ corresponds to the facts $\mathtt{person(1).}$ $\mathtt{person(2).}$ 
In order to formulate cardinality constraints concisely, so called conditional literals are supported. The basic idea is that conditional literals serve as a generator for producing a set of atoms. To exemplify the application of cardinality constrains and conditional literals, let an ASP program contain the facts: 
$\mathtt{thing(3..4).}$ $\mathtt{cabinetDomain(9..10).}$ 
The constraint 
\begin{flalign*}
&\mathtt{1\{cabinetTOthing(X,Y):cabinetDomain(X)\}1} 
                              \leftarrow \mathtt{thing(Y).} &
\end{flalign*}
where $\mathtt{cabinetTOthing(X,Y):cabinetDomain(X)}$ is a conditional literal, is expanded to 
\begin{flalign*}
\mathtt{1\{cabinetTOthing(9,3),}&\mathtt{~cabinetTOthing(10,3)\}1}                          \leftarrow \mathtt{thing(3).} &\\
\mathtt{1\{cabinetTOthing(9,4),}&\mathtt{~cabinetTOthing(10,4)\}1} 
 \leftarrow \mathtt{thing(4).} 
\end{flalign*}
and expresses that things 3 and 4 must be connected to exactly one of the cabinets 9 and 10. Conditional literals can be used in cardinality constraints in place of atoms, where the \emph{conditional part} (e.g.\ $\mathtt{cabinetDomain(X)}$) is a (conjunction of) \emph{domain predicate}(s) preceded by the \emph{main part}. All predicates of atoms in the conditional part must be  domain predicates. Roughly speaking, domain predicates are predicates whose ground instantiations are the same for all answer sets, i.e.\ the ground instantiations can be determined without searching. A variable is local iff it appears only in a conditional literal, e.g.\ $X$ is local in our example. All other variables are global, e.g.\ $Y$. During grounding of the rules, global variables are instantiated first. Then the main part of the conditional literal is expanded for the instantiations of the local variables where the conditional part is fulfilled. 

Answer set programming solvers like~\cite{Simons2002,Leone2006,Gebser2010a} offer optimization services. In particular, the literal $\mathtt{\# minimize[a_1 = w_1,\dots,a_n = w_n]}$ allows minimization, where $\mathtt{a_i}$ are atoms and $\mathtt{w_i}$ are associated weights. An answer set is optimal iff the sum of the weights of atoms which are satisfied in this answer set is minimal among all answer sets of a given program.

\section{Configuration problems}\label{sec:config}

We employ a definition of configuration problems based on logical descriptions~\cite{Soininen2001,Felfernig2004a}. The basic idea is that every finite Herbrand-model contains the description of exactly one configuration. 

The description of a configuration is defined by relations expressed by a set of predicates $\PS$, which is called \emph{solution schema}. 
For our example the solution schema consists of the four unary predicates \texttt{thing/1}, \texttt{person/1}, \texttt{cabinet/1} and \texttt{room/1} representing the individuals and the four binary predicates \texttt{personTOthing/2}, \texttt{personTOroom/2}, \texttt{roomTOcabinet/2} and \texttt{cabinetTOthing/2} representing the relations. 
An instantiation of this solution schema corresponds to a configuration. 

We assume that predicate symbols have a unique arity. The set of Herbrand-models is specified by a set of logical 
sentences $\REQ$, which usually comprises \emph{customer} and \emph{
configuration requirements}. The latter reflect the set of all allowed 
configurations for an artifact, whereas the customer requirements may 
comprise facts and logical sentences specifying the individual needs of 
customers. 
In our example customer requirements are expressed by facts describing the 
relation between persons and things. Configuration requirements, like ``each 
thing must be stored in a cabinet'' are expressed by logical sentences. 

\begin{definition}[Instances of configuration problems]
A configuration problem instance $\langle \REQ, \PS \rangle$ is defined by
 a set of logical sentences $\REQ$ representing requirements and
$\PS$ a set of predicate symbols representing the solution schema. 
For optimization purposes an objective function $f(\Sol) \mapsto \mathbb{N}$  maps any set of atoms $\Sol$ to positive integers where $\Sol$ contains only atoms whose predicate symbols are in $\PS$. 
\end{definition}

Let $\HM(\LS)$ denote the set of Herbrand-models of a set of logical sentences $\LS$ for a given semantics. 

\begin{definition}[Configuration]
$\Sol$ is a configuration for a configuration problem instance $\emph{CPI}=\langle \REQ, \PS \rangle$ iff 
there is a Herbrand-model\/ $\bf M \in \HM(\REQ)$ and $\Sol$ is the set of \emph{all} the elements of\/ $\bf M$ whose predicate symbols are in $\PS$ and $\Sol$ is finite, i.e.\ $\Sol = \{\mathtt{p(\overline{t})} | \mathtt{p} \in {\PS} \mathrm{\ and\ } \mathtt{p(\overline{t})} \in \bf{M})\}$. By $\mathtt{p(\overline{t})}$ we denote a ground instance of $\mathtt{p}$ with a  term vector $\mathtt{\overline{t}}$.
$\Sol$ is an optimal configuration for $\emph{CPI}$ iff $\Sol$ is a configuration for $\emph{CPI}$ and there is no configuration $\Sol '$ of $\emph{CPI}$ s.t.\ $f(\Sol ') < f(\Sol)$.
\end{definition}

\begin{definition}[Configuration generation (optimization) problems]
Let the instances of configuration problems be defined by $\langle \REQ, \PS \rangle$ and objective functions $f(\cdot)$.
Generate a set of atoms $\Sol$ s.t.\ $\Sol$ is a configuration (an optimal 
configuration) for a configuration problem instance.
\end{definition}

The set of Herbrand-models depends on the semantics of the employed logic. In 
this paper, we apply ASP because this framework allows a concise 
and modular specification, assures decidability, and avoids the inclusion of 
unjustified atoms (e.g.\ unjustified components) 
in configurations~\cite{Soininen2001}.

In~\cite{Soininen2001} various modeling patterns based on ASP were introduced. A fixed set of ground facts defines the individuals which are employed for a configuration. 
However, in many cases it is undesirable to consider only a fixed number of individuals used in a configuration. Therefore, we apply the following modeling pattern. 
Let $\mathtt{pLower}$ and $\mathtt{pUpper}$ represent the upper and lower number of individuals of type $\mathtt{p}$. Such a type is called \emph{bounded}.   We require each individual of a configuration, represented by its unique identifier, to be a member of exactly one bounded type. To each bounded type a domain $\mathtt{pDomain}$ is associated, representing the set of possible individuals of the bounded type. We employ numbers as identifiers, starting from some offset. 
For every bounded type $\mathtt{p}$ we add the following axioms:
\vspace{-5pt}\begin{flalign*}
&\mathtt{pDomain(pOffset+1~~..~~pOffset+pUpper).}& 
\mathtt{pLower \{p(X):pDomain(X)\} pUpper.} \qquad \qquad & \\
&\mathtt{p(X) \leftarrow pDomain(X), pDomain(Y), p(Y), X < Y.}&
\vspace{-10pt}\end{flalign*}
\noindent The first rule instantiates the maximal required number of unique individuals of $\mathtt{p}$ in $\mathtt{pDomain}$. The second rule makes sure that at least $\mathtt{pLower}$, but at most $\mathtt{pUpper}$ individuals of $\mathtt{p}$ are asserted. The third rule breaks the symmetry of assertions. 
By these rules the required number of $\mathtt{p}$ individuals are asserted, in order to find a configuration within the given upper and lower bounds.

For some bounded types, e.g. \texttt{person/1} and \texttt{thing/1} the bounds $\mathtt{pLower}$ and $\mathtt{pUpper}$  coincide because the exact number of individuals employed in any configuration is known. In this case the fixed set of $\mathtt{p}$ facts can be asserted without using the rules presented above. 
In our example the customer provides a set of requirements for a configuration including definitions of person and thing individuals as well as their relations: \texttt{person(1..2). thing(3..8). personTOthing(1,3). ... personTOthing(2,8).}
For the bounded type cabinet we add the following rules. The upper and lower numbers of cabinets are computed based on the number of things and persons. The rules for rooms are defined accordingly.
\begin{alltt}\small
cabinetDomain(9..14).      2\{cabinet(X):cabinetDomain(X)\}6.
cabinet(X) :- cabinetDomain(X), cabinetDomain(Y), cabinet(Y), X<Y.
\end{alltt}

Cardinality restrictions given in Section~\ref{sec:example} are encoded with cardinality constraints, where one direction of an association is encoded as a generation rule (see Section~\ref{sec:review}) and the other direction as a constraint. Note, all predicates of atoms in the conditional part of a conditional literal must be domain predicates. 
Therefore, we have to use $\mathtt{pDomain}$ predicates rather than $\mathtt{p}$ predicates, e.g.\ $\mathtt{cabinetDomain(X)}$ instead of $\mathtt{cabinet(X)}$ (see the first rule of the next sequence of rules). However, individuals employed in relations must also be contained in the corresponding types (see the last two rules of the next sequence of rules as example). By these rules we avoid situations where an individual is used in a relation but not included in the bounded type.  

\begin{alltt}\small
1\{cabinetTOthing(X,Y):cabinetDomain(X)\}1 :- thing(Y).
:- 6 \{cabinetTOthing(X,Y):thing(Y)\}, cabinet(X).
cabinet(X) :- cabinetTOthing(X,Y). 
cabinet(Y) :- roomTOcabinet(X,Y).
\end{alltt}

The next rules describe the fact that a room may  contain things of its owner only.
\begin{alltt}\small
personTOroom(P,R) :- personTOthing(P,X), cabinetTOthing(C,X), roomTOcabinet(R,C).
:- personTOroom(P1,R), personTOroom(P2,R), P1!=P2.
\end{alltt}

In addition, optimization can be applied to generate optimal configurations which minimize the overall configuration costs depending on the objective function. 
We model the objective function by assigning to each atom in $\Sol$ some costs. This can be achieved with the following modeling pattern. By the atom $\mathtt{cost(create(a,w))}$, where $a$ is an element of $\Sol$ and $\mathtt{w}$ is an integer, the costs of creating an element $a$ in a configuration are defined. We employ the conjunction of atoms $\mathtt{\alpha(\overline{X},\overline{Y},W)}$ to allow case specific determination of costs. The variable vectors $\mathtt{\overline{X},\overline{Y}}$ can be exploited to formulate arbitrary usage of logical variables in the conjunction of atoms.
For each $\mathtt{p} \in \PS$ include axioms of the following form in $\REQ$:
$\mathtt{cost(create(p(\overline{X})), W) \leftarrow p(\overline{X}), \alpha(\overline{X},\overline{Y},W)}$
such that for each atom $\mathtt{p(\overline{t})}$ in $\Sol$ the answer set contains an atom $\mathtt{cost(create(p(\overline{t}),w))}$ where $\mathtt{w}$ is an integer. For example:
\begin{alltt}\small
roomCost(5). personTOroomCost(1). cost(create(room(X)),W) :- room(X), roomCost(W).
cost(create(personTOroom(X,Y)), W) :- personTOroom(X,Y), personTOroomCost(W).
\end{alltt}
All other creation costs are expressed in the same way. We minimize the sum of all costs by means of the  optimization statement:
$\mathtt{\#minimize[cost(X,W)=W].}$ By this statement all $\mathtt{W}$ of atoms $\mathtt{cost(X,W)}$ are summed up. Only those answer sets which minimize this summation are optimal solutions. 

For the given example the solver finds the optimal configuration including two cabinets and two rooms (depicted in Figure~\ref{figconfig}): 
\texttt{\{cabinet(10), cabinet(9), room(16),} 
\texttt{room(15), \dots,} \\ \texttt{roomTOcabinet(15,9), \dots,} \texttt{cabinetTOthing(10,8), \dots,} \texttt{personTOroom(2,16)\}}

\section{Reconfiguration problems} 

We view reconfiguration as a new configuration-generation problem where parts of a \emph{legacy configuration} are possibly reused. The conditions under which some parts of the legacy configuration can be reused and what the consequences of a reuse are, is expressed by a set of logical sentences $\T$ which relate the legacy configuration $\Sol$ and the new configuration problem instance $\langle \REQ_{\R}, \PR \rangle$. 

\begin{definition}[Instances of reconfiguration problems]
A reconfiguration problem instance \\$\langle \langle \REQ_{\R}, \PR \rangle, \Sol, \T \rangle$ is defined by:
 $\langle \REQ_{\R}, \PR \rangle$ an instance of a configuration problem,
$\Sol$ a legacy configuration and
$\T$ a set of logical sentences representing the transformation constraints regarding the legacy configuration. 
For optimization purposes an objective function $g(\Sol,\R) \mapsto \mathbb{N}$ maps legacy configurations $\Sol$ and configurations $\R$ of $\langle \REQ_{\R}, \PR \rangle$ to positive integers. 

\end{definition}

Note, the two-placed objective function expresses the fact that the costs of a reconfiguration depend on the elements in a reconfiguration and on the reuse or deletion of elements of the legacy configuration. 

In order to avoid name conflicts between the entities of the legacy configuration $\Sol$ and instances of new configuration problems $\langle \REQ_{\R}, \PR \rangle$, we usually formulate $\PR$ and  $\REQ_{\R}$ using  constants not employed in $\Sol$. In particular, we use different name spaces for terms referencing individuals. Together with the unique name assumption this implies that individuals of the legacy configuration and new individuals introduced by the reconfiguration problem are disjunct. 

Reconfigurations are defined analogously to configurations as a finite subset of Herbrand-models. 

\begin{definition}[Reconfiguration]

$\R$ is a reconfiguration for a reconfiguration problem instance $\emph{RCI}=$ $\langle \langle \REQ_{\R}, \PR \rangle, $ $\Sol, \T \rangle$  iff\/ 
there is a Herbrand-model\/ $\bf M \in \HM(\REQ_{\R} \cup \Sol \cup \T)$ and $\R$ is the set of all the elements of\/ $\bf M$ whose predicate symbols are in $\PR$ and $\R$ is finite.
$\R$ is an optimal reconfiguration for $\emph{RCI}$ iff $\R$ is a reconfiguration for $\emph{RCI}$ and there is no reconfiguration $\R '$ of $\emph{RCI}$ s.t.\  $g(\Sol, \R ') < g(\Sol, \R)$.
\end{definition}

\emph{Reconfiguration problems} are formulated analogously to configuration problems. 

\begin{definition}[Reconfiguration generation (optimization) problems]
The instances of reconfiguration problems are defined by a tuple $\langle \langle \REQ_{\R}, \PR \rangle,$ $ \Sol, \T \rangle$ and objective functions $g(\cdot, \cdot)$.
 Generate a set of atoms $\R$ s.t.\ $\R$ is a reconfiguration (an optimal reconfiguration) for a reconfiguration problem instance.

\end{definition}

In the following we show typical formalization patterns and apply them to our example. The set of atoms $\{\mathtt{legacyConfig}(a) | a \in \Sol\}$ describes the atoms of the legacy configuration $\Sol$. Note, the definition of reconfiguration problems does not employ first-order logic constructs in order to avoid unnecessary restrictions. However, to facilitate a concise description of the problem we introduce the predicate $\mathtt{legacyConfig}/1$ to allow quantification over the elements of the legacy configuration. 

For the transformation sentences $\bf T$ we employ the following general patterns.  
For reusing parts of the legacy configuration the problem solver has to make the decision either to \emph{reuse} or to \emph{delete}. This is expressed by $\mathtt{reuse(a)}$ and $\mathtt{delete(a)}$ atoms where $a$ is an element of $\Sol$. For each atom $a$ in $\Sol$ either $\mathtt{reuse(a)}$ or $\mathtt{delete(a)}$ must hold. Based on these atoms additional configuration constraints can be defined which describe the proper reuse or deletion of a part of the legacy configuration represented by atom $a$. In our case, reusing an atom $a$ of the legacy configuration implies the assertion of this atom, whereas deletion requires that the atom is not asserted by any rule application of the knowledge base. In addition, costs are  associated to each $\mathtt{reuse(a)}$ or $\mathtt{delete(a)}$ operation. This is expressed by the atom $\mathtt{cost(reuse(a),w)}$ or $\mathtt{cost(delete(a),w)}$ where $\mathtt{a}$ is an element of $\Sol$ and $\mathtt{w}$ is an integer specifying the corresponding costs. Furthermore, we require that in each model which contains $\mathtt{reuse(a)}$ or $\mathtt{delete(a)}$ also $\mathtt{cost(reuse(a),w)}$ or $\mathtt{cost(delete(a),w)}$ is contained in order to have defined reuse or deletion costs. The conjunctions $\beta(\overline{X},\overline{Y},W)$ and $\gamma(\overline{X},\overline{Y},W)$ are employed to define case specific costs. 
 
For each $\mathtt{p} \in \PS$ include the following axioms in $\T$:
\begin{flalign*}
&\mathtt{1\{reuse(p(\overline{X})), delete(p(\overline{X}))\}1 \leftarrow legacyConfig(p(\overline{X})).}& \\
&\leftarrow \mathtt{p(\overline{X}), delete(p(\overline{X})).} ~~~~
\mathtt{p(\overline{X})\leftarrow reuse(p(\overline{X})).}& \\
&\mathtt{cost(reuse(p(\overline{X})), W) \leftarrow 
\reuse(p(\overline{X})), \beta(\overline{X},\overline{Y},W).} \\
&\mathtt{cost(delete(p(\overline{X})),W) \leftarrow 
delete(p(\overline{X})), \gamma(\overline{X},\overline{Y},W)}.
\end{flalign*}

Analogously to configuration problems, we require each individual contained in a reconfiguration to be a member of exactly one bounded type. Consequently, individuals of the legacy configuration have to be a member of the domain $\mathtt{pDomain(X)}$ of a bounded type $\mathtt{p}$ of $\langle \REQ_{\R}, \PR \rangle$, because these individuals can be part of a reconfiguration through reuse, i.e.\ there are rules of the form 
\begin{flalign*}
&\mathtt{pDomain(X) \leftarrow legacyConfig(q(\ldots,X,\ldots)).}&
\end{flalign*}
where $\mathtt{q}$ is predicate symbol of the solution schema of the legacy configuration. 

As for configuration problems, the number of individuals of a bounded type $\mathtt{p}$ is limited. 
For every bounded type $\mathtt{p}$ we add the following axioms:
\begin{flalign*}
&\mathtt{pLower \{p(X):pDomain(X)\} pUpper.}&
\end{flalign*}
However, the two other rules for bounded types are changed. In particular, we have to adapt the symmetry breaking pattern of configurations. The reason is that there are two different types of individuals contained in $\mathtt{pDomain}$, those which are reused and those which are newly generated. Symmetry breaking does not apply to the reused individuals because they may be linked to other reused individuals. Therefore, exchanging these individuals potentially leads to different configurations. However, the newly generated individuals are interchangeable. We describe them by $\mathtt{pDomainNew/1}$ for the bounded type $\mathtt{p}$. We use $\mathtt{pNewOffset}$ to generate new identifiers, i.e.\ the pattern is: 
\begin{flalign*}
&\mathtt{pDomainNew(pNewOffset+1~~..~~pNewOffset+pUpper).}& \quad
&\mathtt{pDomain(X) \leftarrow pDomainNew(X).} & \\
&\mathtt{p(X) \leftarrow pDomainNew(X), pDomainNew(Y), p(Y), X < Y.}&
\end{flalign*}

In our example, the reconfiguration problem consists of additional customer and configuration requirements described in Section~\ref{sec:example}. The solution schema for the reconfiguration problem is an extension of the solution schema of the original configuration problem by \texttt{cabinetHigh/1}, \texttt{cabinetSmall/1}, \texttt{thingLong/1} and \texttt{thingShort/1} predicates.
The additional customer requirements are expressed by: 
\begin{alltt}\small
thingLong(3). thingShort(4). thingShort(5). thingShort(6). thingShort(7).  
thingLong(8). thing(21).     thingLong(21). personTOthing(1,21).
\end{alltt}
The legacy configuration presented in Section~\ref{sec:config} is encoded using \texttt{legacyConfig} predicate: 
\begin{alltt}\small
legacyConfig(cabinet(9)). legacyConfig(cabinetTOthing(10,8)).  ...
\end{alltt}

To implement the configuration requirements of the modified problem we add rules defining the subtypes of cabinets as well as that long things have to be stored in high cabinets. Note, only some of the usual rules for expressing subtypes are needed. Regarding subtypes of thing, no rules are needed at all because for every \texttt{thing} fact either a \texttt{thingLong} fact or a \texttt{thingShort} fact is contained in the customer requirements and none of these predicates appear in the head of a rule. 
\begin{alltt}\small
1\{cabinetHigh(X), cabinetSmall(X)\}1 :- cabinet(X).
cabinetHigh(C) :- thingLong(X), cabinetTOthing(C,X).
\end{alltt}
Moreover, each high cabinet requires more space in a room. Such a cabinet occupies two of the four available slots in a room, whereas a small cabinet uses only one slot. Note, the fourth rule of the following rule sequence contains a weighted constraint. To each atom $\mathtt{roomTOcabinetSlot(X,Y,S)}$ a weight $\mathtt{S}$ is associated. The weighted constraint is true if the sum of the weights of the true atoms are greater or equal to 5. In this case an inconsistency is detected. 
 
\begin{alltt}\small
cabinetSize(X,1) :- cabinet(X), cabinetSmall(X). 
cabinetSize(X,2) :- cabinet(X), cabinetHigh(X).
roomTOcabinetSlot(R,C,S) :- roomTOcabinet(R,C), cabinetSize(C,S).
:- 5 [roomTOcabinetSlot(X,Y,S):cabinetDomain(Y)=S], room(X).
\end{alltt}
The domain of cabinets is extended with additional individuals that might be required in a new configuration. The number of new elements in the cabinet domain corresponds to the number of things in the modified problem. The upper number \texttt{pUpper} of cabinet individuals is set to 7, because 7 things must be stored in the house. The extension of the room domain is done in the same way.
\begin{alltt}\small
cabinetDomainNew(22..28).   cabinetDomain(X) :- cabinetDomainNew(X).  
2\{cabinet(X):cabinetDomain(X)\}7. 
cabinet(X) :- cabinetDomainNew(X), cabinet(Y), X<Y, cabinetDomainNew(Y).
\end{alltt}

The transformation rules are implemented as described above, e.g.:\
\begin{alltt}\small
1\{reuse(cabinet(X)), delete(cabinet(X))\}1 :- legacyConfig(cabinet(X)).
cabinetDomain(X) :- legacyConfig(cabinet(X)).
\end{alltt}
 However, the transformation rules for \texttt{legacyConfig(person(X))}, \texttt{legacyConfig(thing(X))} as well as for \texttt{legacyConfig(personTOthing(X,Y))} could be deleted because facts about persons, things and their relations are given as requirements. Deleting such an atom results in a contradiction. 

Given the reconfiguration program the solver identifies a reconfiguration as well as a set of actions required to transform the legacy configuration into a new one. 
For generating optimal reconfigurations we formulate a cost model. The minimization statement in the reconfiguration problem is the same as in the configuration. In our reconfiguration example the costs for creation of new high/small cabinets and rooms $\mathtt{cost(create(a),w)}$ correspond to the costs definition of the configuration problem.
To obtain a reconfiguration scenario with the minimal costs of required actions we extend the costs rules described above with costs for creation of new high/small cabinet and room individuals as well as with costs for newly created relations, e.g.:\ 
\begin{alltt}\small
cost(create(cabinetHigh(X)),W):-cabinetHigh(X),cabinetHighCost(W),cabinetDomainNew(X).
\end{alltt}
Rules for deducing the costs of reuse and deletion are formulated as described above. 

For our example let us assume that the customer sets all deletion costs to 2, whereas reusing has no costs except for cabinets, which could be altered to high in a reconfiguration. The costs of this alteration is set to 3. Creation costs of new high and small cabinets are set to 10 and 5 respectively. Finally, the costs of a new room is set to 5. Creation of relations between individuals is for free. Given these costs the solver is able to find a set of optimal reconfigurations including the one presented in Figure~\ref{figreconfigsol1}.

Modification of the costs results in different optimal reconfigurations. 
Let us assume the sales-department changes both the costs of deletion of a 
cabinet and the costs of increasing the height of a cabinet to 10, and 
decreases the creation costs of new high and small cabinets to 2 and 1 
respectively. In this case the solutions returned by a solver will include 
the one presented in Figure~\ref{figreconfigsol2}. 
Given their simplicity, the presented optimal solutions 
were found in milliseconds.

\section{Evaluation}
We are challenged by various real-world configuration problems from technical domains like telecommunication or railway safety systems with structure and complexity similar to the example used in this paper. The evaluation of our approach is performed on test cases\footnote{Available from: http://proserver3-iwas.uni-klu.ac.at/reconcile/index.php/benchmarks} from such practical configuration scenarios. Each scenario can be represented as an instance of the (re)configuration problem presented in Section~\ref{sec:example}. In the  \emph{empty} reconfiguration scenario the legacy configuration is empty and the customer requirements contain sets of things and persons owning 5 things each. Things are labeled as short. The reconfiguration process should create missing cabinets, rooms and relations. 

The customer requirements of the \emph{long} scenario specify that each given person owns 15 things. The legacy configuration contains a set of relations that indicate placement of these things into cabinets, s.t. all things of one person are stored in three cabinets that are placed in one room. The customer also requires 5 things of each person to be labeled as long whereas the remaining 10 as short. The goal of the reconfiguration is to find a valid rearrangement of long things to reused or newly created high cabinets. 

The next \emph{new room} scenario models a situation when new rooms have to be created and some of the cabinets reallocated. In this scenario each person owns 12 things. These things are stored in 3 cabinets placed in one room as indicated by the legacy configuration. In the reconfiguration problem the customer requirements declare 6 of the 12 things as long. 

The last scenario, \emph{swap}, describes a situation when the customer requirements include only one person, who owns 35 things. In the legacy configuration the things are placed in 3 cabinets in the first room and in 4 cabinets in the second room. Moreover, one of the things in the second room is labeled as long in the customer requirements. Given the costs schema presented above, the solution corresponds to a rearrangement of the cabinets in the rooms such that a high cabinet can be placed in one of these rooms. 
All these scenarios can be easily scaled by increasing the number of things. The number of persons in the empty, long and new room scenarios can always be computed given the number of things.

\begin{figure}
	\centering
		\includegraphics[width=12cm]{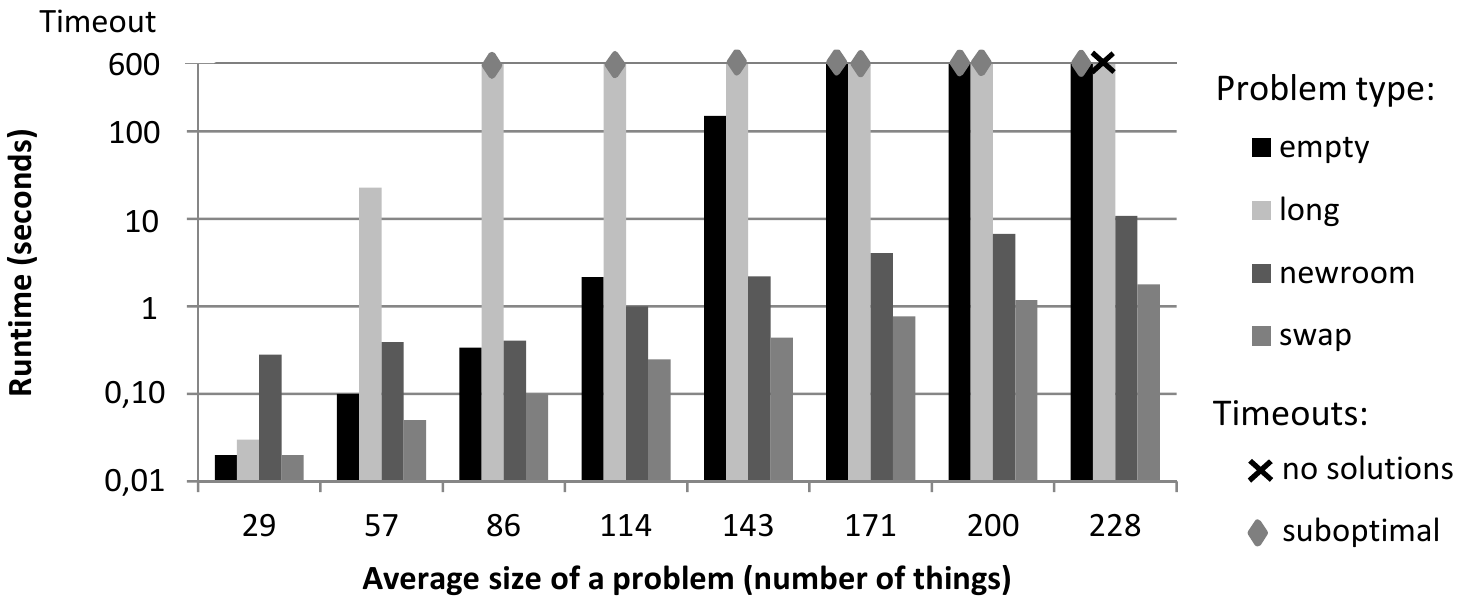}
	\caption{Evaluation results}
	\label{fig:evaluation}
	\vspace{-15pt}
\end{figure}

Experiments were performed using Potassco 3.0.3 on Core2 Duo 3Ghz with 4Gb RAM. In our experiments we considered only creation costs for newly generated cabinets and rooms because these are the dominant costs for our application domain. The performance of the reconfiguration process is presented in Figure~\ref{fig:evaluation}. Potassco was able to find optimal solutions within 600 seconds for all instances of the new room and swap scenarios. Optimal solutions were also found for small and mid-size instances of the empty scenario. For all other instances at least one suboptimal solution was found. The long scenario included the hardest problems. The solver did not find any solutions for one of them in 600 seconds. This was the only unsolved problem instance in the whole experiment. Because the solved instances are comparable to real world applications based on our experiences, we consider the proposed reconfiguration method as feasible for a practically interesting set of reconfiguration problem instances.

\section{Conclusions and related work}

The existing approaches for reconfiguration can be separated into revision-based \cite{Mannisto1999} and model-based \cite{Stumptner1998a}. The revision-based approaches employ a knowledge base describing ``fixes'', i.e.\ reconfiguration operations and configuration invariants~\cite{Mannisto1999}. A solution requires that there is a \emph{sequence} of operations which transform the legacy configuration into a new configuration. The approach of \cite{Stumptner1998a} views reconfiguration as a consistency-maintenance (diagnosis) problem, where a solution corresponds to a consistent set of assumptions s.t.\ requirements are implied. Similarly, our approach can be seen as searching for a consistent (optimal) set of assumptions regarding reuse or deletion of parts of the legacy configuration and creation of new parts. This search is provided by the ASP reasoning system, implementing a \emph{correct and complete} problem solving method. No additional diagnosis component is required. Regarding the revision-based approach, our domains do not need the computation of sequences of operations, because if a reconfiguration is found, a sequence of real-world change operations can be easily derived. Thus, we can avoid the additional combinatorial explosion introduced by permutations of change operations.  However, we can view our approach as a form of the revision-based method assuming that all change operations are executed simultaneously. The effects of these operations and the combination of allowed operations are described by the transformation knowledge. Thus we can model complex ``fix'' operations which involve the reuse of several parts of the legacy configuration and which have multiple effects such as creating new parts or deleting existing ones. 

To sum up, we have developed a method which allows the modeling of reconfiguration problems based on legacy configurations, transformation knowledge, and a new configuration problem instance. We showed various modeling patterns and implemented the approach based on ASP. Evaluation results show the feasibility for practical applications. 

\bibliographystyle{eptcs}
\bibliography{reconfig}

\end{document}